\documentclass[10pt,twocolumn,letterpaper]{article}

\ifdefined\pdfminorversion
\fi

\usepackage[pagenumbers]{cvpr} %

\usepackage{multirow}
\usepackage{placeins}
\usepackage[T1]{fontenc}
\usepackage{tcolorbox}
\tcbuselibrary{skins,breakable}
\newcommand{\findingblock}[2]{%
\par\begingroup
\ifdefined\nolinenumbers\nolinenumbers\fi
\begin{tcolorbox}[
  enhanced jigsaw,
  breakable,
  lines before break=3,
  colback=blue!4!white,
  colframe=blue!30!gray!90!black,
  borderline west={1.5pt}{0pt}{blue!30!gray!90!black},
  boxrule=0pt,
  sharp corners,
  boxsep=0pt,
  left=6pt, right=6pt, top=3pt, bottom=4pt,
  before skip=6pt, after skip=6pt,
  pad at break*=3pt,
  title={#1},
  fonttitle=\small\bfseries,
  coltitle=blue!30!gray!90!black,
  colbacktitle=blue!4!white,
  titlerule=0pt,
  lefttitle=6pt, righttitle=6pt, toptitle=4pt, bottomtitle=5pt,
  halign title=flush left,
  fontupper=\small,
]
\begin{itemize}[leftmargin=10pt, labelsep=4pt, itemsep=2pt,
  topsep=0pt, parsep=0pt, partopsep=0pt]
\interlinepenalty=10000
#2
\end{itemize}
\end{tcolorbox}\par
\endgroup
}

\definecolor{cvprblue}{rgb}{0.21,0.49,0.74}
\usepackage[pagebackref,breaklinks,colorlinks,allcolors=cvprblue]{hyperref}

\def\confName{CVPR}
\def\confYear{2027}

\title{SpatialToM: Modeling Multi-Agent Theory of Mind \\for Embodied Spatial Intelligence}

\author{
Han Yin$^{1,*}$ \quad
Yining Hong$^{1,*}$ \quad
Jiageng Liu$^{2,*}$ \quad
Jieneng Chen$^{1}$ \quad
Ruohan Zhang$^{3}$ \quad
Manling Li$^{3}$\\[0.25em]
Li Fei-Fei$^{1}$ \quad
Jiajun Wu$^{1}$ \quad
Tianmin Shu$^{4,\dagger}$ \quad
Leonidas Guibas$^{1,\dagger}$ \quad
Yejin Choi$^{1,\dagger}$\\[0.5em]
{\normalsize
$^{1}$Stanford University \qquad
$^{2}$UCLA \qquad
$^{3}$Northwestern University \qquad
$^{4}$Johns Hopkins University
}\\[0.3em]
{\small $^{*}$Equal contribution. \quad $^{\dagger}$Co-advising.}\\[0.35em]
{\small\url{https://spatialtom-bench.github.io/}}
}

\begin{document}
\raggedbottom

\onecolumn
\begingroup
\makeatletter
\patchcmd{\@maketitle}{\vskip .375in}{\vspace*{-18pt}}{}{
  \PackageError{spatialtom}{Could not adjust title inset}{Check the CVPR title definition.}}
\patchcmd{\@maketitle}{\Large \bf \@title \par}{%
  \fontsize{18}{22}\selectfont\bfseries \@title \par}{}{
  \PackageError{spatialtom}{Could not adjust title size}{Check the CVPR title definition.}}
\makeatother
\maketitle
\endgroup
\vspace{-2.5em}
\begin{center}
    \includegraphics[width=\textwidth]{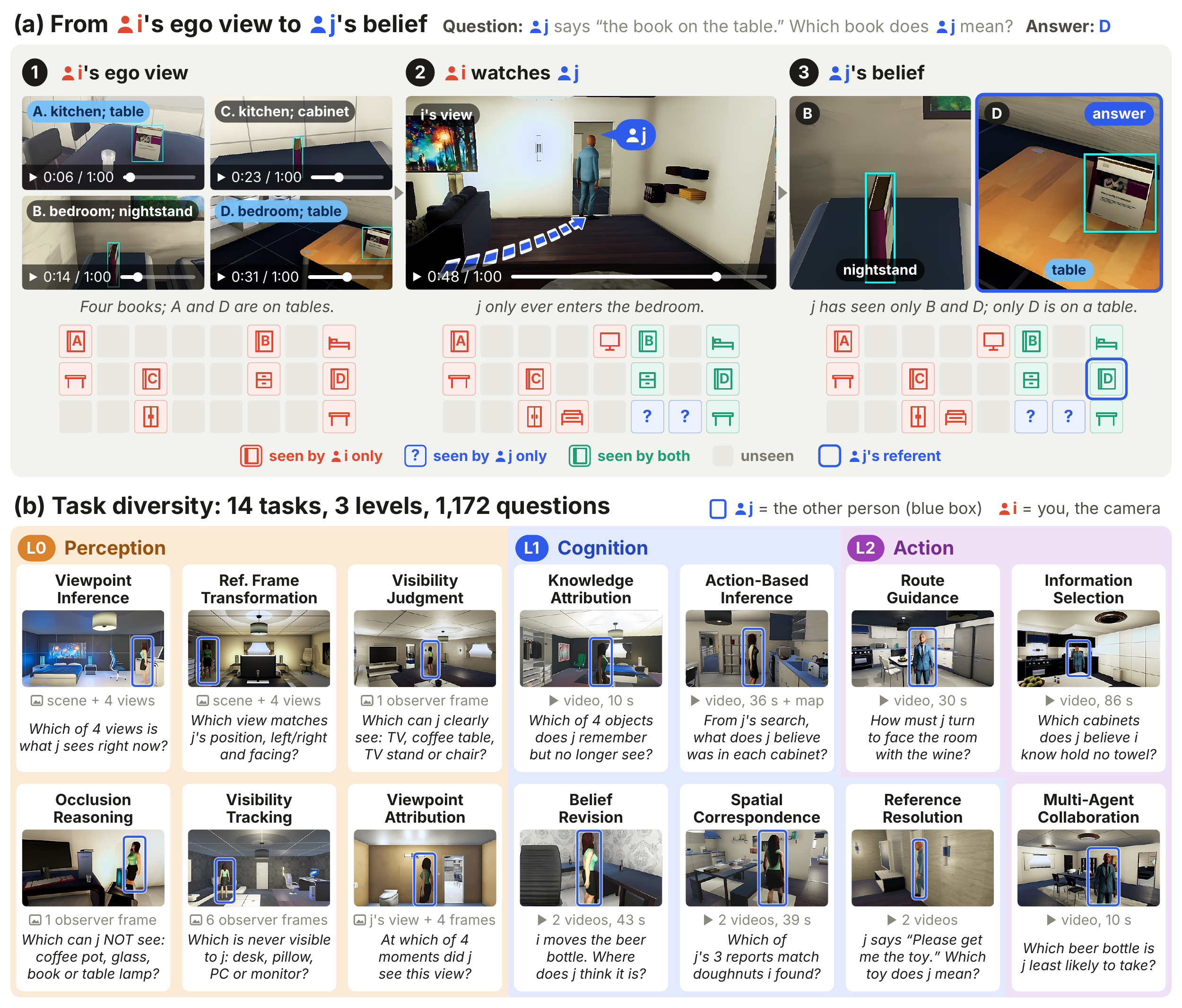}
    \vspace{-1.5em}
    \captionsetup{hypcap=false}
    \captionof{figure}{Overview of SpatialToM.
    \textbf{(a)} The observer has seen all four books,
    but $j$ has seen only two; resolving $j$'s request
    for ``the book on the table'' requires modeling what
    $j$ has seen.
    \textbf{(b)} The 14 task types across Perception (L0),
    Cognition (L1), and Action (L2).}
    \label{fig:teaser}
\end{center}
\twocolumn

\begin{abstract}

Spatial intelligence is not merely egocentric, but fundamentally intersubjective: it emerges where our perceived world and the unseen worlds of other agents seamlessly interlock.  We introduce \textbf{SpatialToM}, a benchmark for embodied spatial theory of mind, where agents must reconstruct others' mental states from spatial perspectives, observation histories, and goal-directed actions. SpatialToM organizes 14 task types into three hierarchical tiers: \textbf{(L0) Perception}, reasoning about perspective and visibility; \textbf{(L1) Cognition}, attributing and integrating others' knowledge and beliefs; and \textbf{(L2) Action}, leveraging these inferences for guidance and multi-agent coordination. Built in VirtualHome, SpatialToM comprises 1,172 questions grounded in agents' distinct perceptual, cognitive and active experiences.
Evaluation of state-of-the-art VLMs reveals that
explicit memory and denser observations offer selective gains
in spatial grounding and knowledge attribution, but remain
vulnerable to perceptual errors and belief reconstruction failures.
Oracle physical records improve accuracy
from 61.1\% to 87.1\%, yet models still struggle with belief
uncertainty, spatial-temporal reasoning, and collaborative actions. These findings expose
a gap between perceiving the world
and understanding the minds that inhabit it. \looseness=-1
\end{abstract}

\section{Introduction}
\label{sec:intro}

Human spatial awareness effortlessly transcends our individual retinas; it emerges through the interlocking of our experiences and those of others \citep{merleauponty2012phenomenology}. Spatial intelligence is thus not merely egocentric, but fundamentally \textit{intersubjective}. Watch a partner open an opaque cabinet, pause, and walk away empty-handed: even with the interior blocked from your view, you instantly infer the target is likely not inside.  In that moment, another actor ceases to be a mere physical obstacle and becomes an epistemic lens into the occluded world. We do not merely track bodies in space; we track minds in space, weaving our own sensory history with the inferred perceptions and beliefs of others to construct a richer understanding of the shared spatial world.

This synthesis of spatial perception and social mentalizing, which we term \textbf{\textit{Embodied Spatial Theory-of-Mind}}, is a cornerstone of human spatial intelligence, yet it remains underexplored in modern AI systems. Spatial intelligence and theory of mind have often been studied separately, leaving their interaction under partial observability insufficiently understood. On one hand, embodied spatial reasoning~\citep{azuma2022scanqa,ma2023sqa3d,hong2023threedllm,yang2025thinking} predominantly treats the environment as a solitary arena. On the other hand, many existing Theory-of-Mind (ToM) benchmarks~\citep{le2019revisiting,gandhi2023understanding,he2023hitom,kim2023fantombenchmarkstresstestingmachine,xu2024opentom} reduce mentalizing to disembodied, symbolic logic puzzles or abstract language reasoning, overlooking perception and embodied interaction as fundamental mechanisms through which we understand what others know and through which ToM itself emerges~\citep{baroncohen1995mindblindness}. Recent embodied ToM benchmarks~\citep{jin2024mmtom,shi2025muma,fan2025somi,juneja2026enacttom} move closer to physical interaction, yet their knowledge asymmetries are largely specified by task design or inferred from behavior, rather than arising from spatial perception itself.

To bridge this gap, we introduce SpatialToM, a benchmark and evaluation framework designed to test embodied spatial theory of mind under partial observability. As illustrated in Figure~\ref{fig:teaser}(a), understanding another agent's request for ``the book on the table'' requires inferring which books that agent has seen, not only which books the observer knows.
Inspired by cognitive accounts of theory of mind~\cite{baroncohen1995mindblindness}, we organize the benchmark into three hierarchical capability tiers, each building on the previous as a foundation for increasingly sophisticated embodied theory-of-mind reasoning: from understanding another agent's perceptual access, to attributing its knowledge and beliefs, and ultimately to leveraging these inferences for guidance and coordination, toward more sophisticated embodied theory of mind.\looseness=-1

We operationalize this three-tier hierarchy through 14 distinct task types (Figure~\ref{fig:teaser}(b)): (i) \textbf{Perception: Perspective and Visibility (L0)}, establishing who can see what and from whose coordinate frame; (ii) \textbf{Cognition: Belief Attribution and Integration (L1)}, attributing knowledge to the other agent from its observations and actions, tracking belief revision, and integrating this knowledge with the observer's own experience through spatial correspondence and reference resolution; and (iii) \textbf{Action: Guidance and Coordination (L2)}, using these inferences to guide collaborative actions in route guidance, information selection, and multi-agent collaboration tasks. To systematically evaluate these capabilities under partial observability, we construct SpatialToM in VirtualHome, comprising 1,172 questions across the 14 task types. Each question is grounded in agents' distinct viewpoints and observation histories, with ground-truth answers derived from their spatial experiences and answerability verified against the observer's available evidence. \looseness=-1

We evaluate current VLMs' spatial theory-of-mind capabilities through four studies. First, we compare direct inference with narrative and graph memories, finding model- and task-dependent gains in spatial grounding and knowledge attribution. Second, denser visual observations improve spatial correspondence but offer limited benefits for belief reconstruction. Third, oracle physical world records substantially improve performance, yet models still struggle with belief uncertainty and spatial-temporal reasoning. Finally, a qualitative planning study reveals a gap between explicitly prompted knowledge attribution and its spontaneous use in multi-agent coordination.

\section{Related Works}
\label{sec:related}

\begin{table*}[t]
\centering
\captionsetup{skip=4pt}
\caption{\textbf{Comparison with representative spatial and ToM benchmarks.}
\textbf{Spatial}: visual reasoning about geometry, reference frames, or viewpoint-dependent visibility.
\textbf{Belief}: others' knowledge, beliefs, or awareness.
\textbf{Coop.}: questions in multi-agent assistance, coordination, or shared-goal scenarios, including VQA.
\textbf{Spatial grounding}: structured spatial information used in task construction.
\ensuremath{\checkmark}: dedicated coverage; ---: otherwise.
I/V/T: image(s)/video/text; Act.: interactive.
$\dagger$ESI-Bench also tests proximity-based agent exposure.}
\label{tab:benchmark_comparison}
\begingroup
\fontsize{8.5}{9.7}\selectfont
\setlength{\tabcolsep}{3pt}
\renewcommand{\arraystretch}{1.0}
\newcommand{\bYes}{\ensuremath{\checkmark}}
\newcommand{\bNo}{---}
\begin{tabular*}{.90\textwidth}{@{\extracolsep{\fill}}lllccc@{}}
\toprule
\textbf{Benchmark} & \textbf{Format} & \textbf{Spatial grounding} & \textbf{Spatial} & \textbf{Belief} & \textbf{Coop.} \\
\midrule
VSI-Bench~\citep{yang2025thinking} & V / QA & 3D scene annotations & \bYes & \bNo & \bNo \\
3DSRBench~\citep{ma2025threedrsbench} & I / QA & Human spatial annotations & \bYes & \bNo & \bNo \\
MindCube~\citep{wang2026mindcubespatialmentalmodeling} & I / QA & Multi-view spatial annotations & \bYes & \bNo & \bNo \\
ESI-Bench~\citep{hong2026esibench} & Act.\ / QA & Sim.\ geometry + object states & \bYes & \bNo$^\dagger$ & \bNo \\
Theory of Space~\citep{zhang2026theoryspacefoundationmodels} & Act.\ / QA, map & Sim.\ coordinates + pose/FOV & \bYes & \bNo & \bNo \\
FANToM~\citep{kim2023fantombenchmarkstresstestingmachine} & T / QA & --- & \bNo & \bYes & \bNo \\
SimpleToM~\citep{gu2026simpletomexposinggapexplicit} & T / QA & --- & \bNo & \bYes & \bNo \\
MMToM-QA~\citep{jin2024mmtom} & V+T / QA & Symbolic locations + access rules & \bNo & \bYes & \bNo \\
MuMA-ToM~\citep{shi2025muma} & V+T / QA & Symbolic locations + access rules & \bNo & \bYes & \bYes \\
SoMi-ToM~\citep{fan2025somi} & I/V+T / QA & --- & \bNo & \bYes & \bYes \\
AGENT+ (POPKI)~\citep{10.1162/OPMI.a.375} & V / rating & 3D occlusion geometry & \bYes & \bYes & \bNo \\
\midrule
\textbf{SpatialToM (ours)} & \textbf{I/V / QA} & \textbf{Sim.\ geometry + camera/depth} & \bYes & \bYes & \bYes \\
\bottomrule
\end{tabular*}
\endgroup
\vspace{-1.5em}
\end{table*}

Recent advances in \textbf{spatial intelligence} ground language in 3D representations~\citep{hong2023threedllm,gu2023conceptgraphs,huang2024leo,Hong_2024_CVPR} and strengthen geometric reasoning in vision-language models~\citep{chen2024spatialvlm,cheng2024spatialrgpt,ma2025spatialllmcompound3dinformeddesign,song2026robospatialteachingspatialunderstanding,zhang2026flatlandspaceteachingvisionlanguage,cai2025spatialbotprecisespatialunderstanding}. Benchmarks such as VSI-Bench and 3DSRBench expose spatial reasoning limitations~\citep{yang2025thinking,ma2025threedrsbench,du2024embspatialbenchbenchmarkingspatialunderstanding,OpenEQA2023,jia2026omnispatialcomprehensivespatialreasoning}, while ViewSpatial-Bench and MindCube probe perspective-taking~\citep{li2025viewspatial,wang2026mindcubespatialmentalmodeling}. Most closely related, ESI-Bench recasts the observer as an actor, evaluating how agents explore, manipulate, and accumulate evidence to resolve otherwise unobservable spatial properties~\citep{hong2026esibench}. Theory of Space probes the explorer's \emph{own} spatial belief construction and revision~\citep{zhang2026theoryspacefoundationmodels}. SpatialToM instead targets the \emph{intersubjective dimension} of spatial intelligence: reconstructing \emph{others' spatial knowledge} from inferred perceptions and observed actions, distinguishing current visual access from retained knowledge.

In cognitive science, \textbf{Theory of Mind} (ToM) concerns attributing mental states to others~\citep{premack1978does,wimmer1983beliefs}, with visual perspective-taking~\citep{flavell1977development} and inverse planning~\citep{baker2009action} connecting perception and behavior. Benchmarks including ToMi, BigToM, Hi-ToM, FANToM, OpenToM, ToMBench, and ToMATO test false beliefs, recursive attribution, and information asymmetry~\citep{le2019revisiting,gandhi2023understanding,he2023hitom,kim2023fantombenchmarkstresstestingmachine,xu2024opentom,chen2024tombenchbenchmarkingtheorymind,shinoda2025tomatoverbalizingmentalstates}; adversarial evaluations stress-test reasoning robustness~\citep{sclar2024exploretheorymindprogramguided,shapira-etal-2024-clever}. SimpleToM distinguishes mental-state inference from behavior prediction and judgment~\citep{gu2026simpletomexposinggapexplicit}. SimToM filters textual context by character knowledge~\citep{wilf2023thinktwiceperspectivetakingimproves}, while PercepToM separates perception inference from perception-to-belief inference over textual events and conversations~\citep{jung2024perceptionsbeliefsexploringprecursory}. SpatialToM instead requires inferring another agent's visual belief and spatial world model from viewpoint, occlusion, and observation history, making spatial perception itself part of knowledge and belief attribution.

Research on \textbf{embodied and multi-agent social reasoning} brings these questions closer to physical interaction. Tang et al.~\citep{tang-etal-2024-grounding} study multi-perspective referential communication in 3D environments. MMToM-QA and MuMA-ToM evaluate object-location belief and goal inference from household activities~\citep{jin2024mmtom,shi2025muma}; SoMi-ToM evaluates ToM from first- and third-person observations~\citep{fan2025somi}; and ToM-Video grounds mental-state inference in narrative-video evidence~\citep{liu-etal-2026-tom}. EnactToM evaluates belief-guided coordination~\citep{juneja2026enacttom}, and MindPower tests false-belief correction and goal-based assistance~\citep{zhang2026mindpowerenablingtheoryofmindreasoning}. ToM-SSI studies belief inference under spatially constrained communication~\citep{bortoletto2025tomssievaluatingtheorymind}, and POPKI models how 3D occlusion and perceptual history shape knowledge and surprise~\citep{10.1162/OPMI.a.375}. Meanwhile, VirtualHome and BEHAVIOR-1K provide situated environments~\citep{puig2018virtualhome,li2023behavior}; Watch-and-Help, PARTNR, CoELA, CHAIC, and Smart Help study assistance and collaborative planning~\citep{puig2021watchandhelpchallengesocialperception,chang2025partnr,ICLR2024_54b8b4e0,NEURIPS2024_4eb8e997,cao2024smarthelpstrategicopponent}. SpatialToM treats \emph{others' spatial knowledge} as both a target of \emph{belief attribution} and a source of evidence about shared space. Grounded in agents' viewpoints and observation histories, it connects knowledge attribution and belief revision with \emph{cross-agent spatial integration}, \emph{guidance}, and \emph{coordination}, using others as an \emph{epistemic lens} to infer unseen information, resolve spatial references, and coordinate complementary actions.

Table~\ref{tab:benchmark_comparison} compares benchmark formats and spatial grounding, alongside spatial reasoning, belief attribution, and cooperation-oriented questions.

\section{SpatialToM Benchmark}
\label{sec:benchmark}
\begin{figure*}[htbp]
    \centering
    \includegraphics[width=\textwidth]{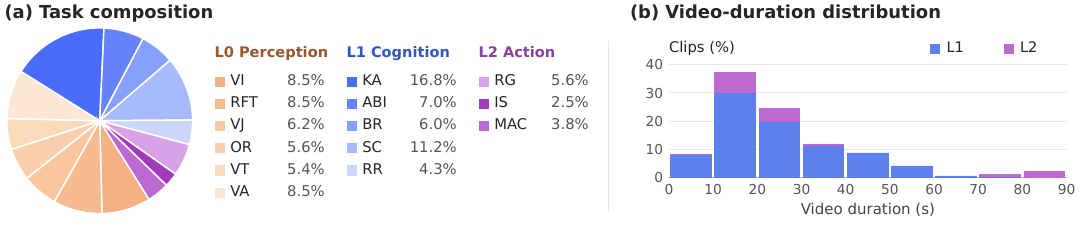}
    \vspace{-2em}
    \caption{Dataset statistics. Left: question proportions by task type, grouped by Perception (L0), Cognition (L1), and Action (L2). Right: shares of all unique video clips in 10-second duration bins, excluding static images and image sequences.}
    \label{fig:dataset_statistics}
    \vspace{-1em}
\end{figure*}

In this section, we introduce SpatialToM, a benchmark for evaluating embodied spatial theory of mind in 3D environments.
Inspired by cognitive accounts of visual perspective-taking~\citep{flavell1977development}, mental-state attribution~\citep{Baker_2017}, and joint action~\citep{SEBANZ200670}, SpatialToM organizes 14 task types into three progressive tiers: \textbf{Perception}, \textbf{Cognition}, and \textbf{Action}.
The benchmark contains 1,172 questions, with visual inputs comprising static images, ordered image sequences, and 812 unique video clips. Figure~\ref{fig:dataset_statistics} summarizes the task composition and video-duration distribution.

\subsection{Benchmark Setup}
\label{sec:setup}

\paragraph{Task definition and notation.}
SpatialToM evaluates visual question answering grounded in the egocentric observations of agent $i$, who explores the environment and observes another agent $j$.
Different viewpoints and exploration histories create asymmetries in spatial knowledge; the model uses $i$'s visual observations and the stated context to reason about $j$'s visual access, knowledge, and behavior.
We denote $i$'s spatial knowledge and beliefs by $B_i$, its observation history, including event order and timing, by $H_i$, and $j$'s actions observed within that history by $A_{j \mid i}$.
The agents' goals, when specified, are $g_i$ and $g_j$.

\noindent\textbf{Task format.}
Each instance pairs visual observations with a question and reference answer; additional inputs may include reference or candidate views, goals, conditions, or utterances.
All tasks, including action judgments, use visual question answering with a single option or a set of option labels or object identifiers.

\subsection{Capability and Task Taxonomy}
\label{sec:task_taxonomy}
The three tiers connect visual access, knowledge attribution and integration, and knowledge use in interaction; Figure~\ref{fig:teaser}(b) illustrates the 14 task types.

\subsubsection{Perception: Perspective and Visibility}
\label{sec:perception}

Attributing spatial knowledge starts with reconstructing the visual evidence available from another viewpoint.
The Perception tier (L0) evaluates this spatial foundation: $i$ uses scene layout and $j$'s position, orientation, and motion to infer its visual access under field-of-view and occlusion constraints.
Six image-based tasks cover perspective transformation, object visibility, and viewpoint tracking, connecting the geometry of individual views with consistency across observations:
\begin{equation}
    \label{eq:perspective_inference}
    (B_i, H_i, A_{j \mid i})
    \longrightarrow \hat{O}_{j \mid i},
\end{equation}
where $\hat{O}_{j \mid i}$ denotes $i$'s estimate of $j$'s visual observations at the relevant moments.

\textit{Perspective transformation} tests how spatial information is interpreted from $j$'s viewpoint: \textbf{Viewpoint Inference (VI)} distinguishes the actual view from alternatives with shifted positions or orientations, while \textbf{Reference Frame Transformation (RFT)} tests left--right and directional consistency using mirrored images or alternative viewpoints.
\textit{Visibility attribution} tests object-level visual access under occlusion and field-of-view constraints: \textbf{Visibility Judgment (VJ)} identifies a visible object among invisible alternatives, while \textbf{Occlusion Reasoning (OR)} identifies an occluded object among visible alternatives, or a visible object among occluded or out-of-view alternatives.
\textit{Perspective tracking} connects observations across time: \textbf{Visibility Tracking (VT)} identifies an object never visible to $j$ at any shown moment, while \textbf{Viewpoint Attribution (VA)} matches a given first-person view to the moment with the corresponding position and orientation.

\subsubsection{Cognition: Belief Attribution and Integration}
\label{sec:cognition}

Spatial knowledge depends on accumulated experience, including information retained outside the current view and uncertainty about subsequent changes.
The Cognition tier (L1) tests both the attribution of this knowledge to $j$ and its integration with $i$'s own experience.
This grouping treats another agent's knowledge as both a target of inference and a source of evidence for understanding shared space.

\vspace{-0.5em}
\paragraph{Spatial Belief Attribution.}
Observation histories support forward inference from visual access to knowledge, while goal-directed actions support backward inference about what the actor believes:
\begin{equation}
    \label{eq:spatial_belief_attribution}
    (B_i, H_i, \hat{O}_{j \mid i}, A_{j \mid i}, g_j)
    \longrightarrow B_{ij},
\end{equation}
where $B_{ij}$ denotes $i$'s estimate of $j$'s spatial knowledge and beliefs, including uncertainty.

\noindent\textbf{Knowledge Attribution (KA).}
Reconstruct what $j$ has observed and in what order, distinguishing current visual access from knowledge acquired and retained through earlier observations.

\noindent\textbf{Action-Based Inference (ABI).}
Infer $j$'s beliefs about whether locations not directly visible to $i$ initially contained the target, using $j$'s search and collection behavior under a stated goal.

\noindent\textbf{Belief Revision (BR).}
Use $j$'s observation history to infer what it sees of an object transfer by $i$ and whether it can identify the moved object among same-category alternatives.
These jointly determine whether its identity and location beliefs are retained, revised, or remain uncertain.

\vspace{-0.5em}
\paragraph{Cross-Agent Spatial Integration.}
Cross-agent spatial integration relates another agent's knowledge to one's own experience by aligning spatial context across partial observations.
This alignment helps determine whether $j$'s observations and reports concern objects and locations already encountered by $i$ or parts of the environment outside $i$'s prior experience.
Interpreting $j$'s exploration and utterances through $i$'s spatial memory constrains $j$'s possible beliefs and establishes object correspondences or resolves references: \looseness=-3
\begin{equation}
    \label{eq:cross_agent_integration}
    (B_i, H_i, A_{j \mid i}, u_{j \rightarrow i})
    \longrightarrow (B_{ij}, C_{ij}),
\end{equation}
where $u_{j \rightarrow i}$ denotes $j$'s utterance and $C_{ij}$ the resulting object correspondences or resolved referent.

\noindent\textbf{Spatial Correspondence (SC).}
Identify which of $j$'s reports refer to objects previously encountered by $i$, aligning spatial context across separate experiences to distinguish shared instances from similar objects.

\noindent\textbf{Reference Resolution (RR).}
Resolve $j$'s reference between two same-category objects known to $i$ using distances, connectivity, elapsed time, and estimated exploration pace to constrain its possible observations during partly hidden exploration.
Retain uncertainty when both referents remain possible. \looseness=-1

\subsubsection{Action: Guidance and Coordination}
\label{sec:action}

The Action tier (L2) tests how attributed knowledge constrains guidance and coordination.
We distinguish three requirements: guidance must fit the recipient's spatial frame and knowledge, information selection must reflect the partner's beliefs about the observer's knowledge, and complementary actions must account for the partner's expected choice.
This connects reference-frame reasoning, second-order knowledge attribution, and spatial decision-making within a shared interaction:
\begin{equation}
    \label{eq:guidance_coordination}
    (B_i, B_{ij}, H_i, g_i, g_j)
    \longrightarrow
    B_{iji}\ \text{or}\ a\ \text{or}\ u_{i \rightarrow j},
\end{equation}
where $B_{iji}$ denotes $i$'s estimate of what $j$ believes $i$ knows, $a$ an action selection or prediction, and $u_{i \rightarrow j}$ an instruction addressed to $j$.

\noindent\textbf{Route Guidance (RG).}
Select guidance or predict turns in $j$'s reference frame by aligning its pose with remembered locations and room connectivity.
For turns toward a previously observed location, preserve $j$'s earlier spatial knowledge despite subsequent changes known to $i$.

\noindent\textbf{Information Selection (IS).}
Identify which inspected locations $j$ believes $i$ knows contain no target, using observation history and $j$'s visual access to $i$.
This second-order attribution distinguishes $i$'s privately acquired knowledge from knowledge $j$ attributes to it through observed inspections, providing a basis for selecting nonredundant information during coordination.

\noindent\textbf{Multi-Agent Collaboration (MAC).}
Predict $j$'s choice from its observation history and stated selection rule, then use relative distances to select a complementary object unobserved by $j$.
The agents choose simultaneously without communication.

\subsection{Dataset Construction}
\label{sec:dataset_construction}
We construct SpatialToM in VirtualHome~\cite{puig2018virtualhome}, varying object arrangements, viewpoints, and exploration and interaction sequences to create asymmetries in visual access and accumulated spatial knowledge.
The pipeline connects scene and behavior generation with rendered evidence, reference-answer computation, and verification from $i$'s perspective.

\noindent\textbf{Scene and object selection.}
We select focal objects, containers or surfaces, and landmarks using scene-graph categories, geometry, room membership, spatial relations, and supported interactions.
Placements specify support or containment relations; initial agent positions are screened against scene geometry, with viewing directions set relative to objects or the other agent.
For example, BR varies layouts and viewpoints to change visual access to transfers and evidence for distinguishing same-category objects.
Simulator-returned states and action outcomes are used to verify placements and interactions.

\noindent\textbf{Route and interaction planning.}
Procedural sampling and language-model proposals specify destinations, visit order, viewing directions, and interactions under layout, action, and observation constraints.
For example, KA routes vary encounter order and subsequent visibility to distinguish retained knowledge from current visual access.
After destination and action--object checks, proposals are translated into native navigation and interaction commands.
We check whether executed routes establish the required observations using simulator states, recorded motion, and measured visibility.
Failure summaries feed back to the route planner for revision; remaining invalid candidates are excluded.
\label{sec:experiments}
\begin{figure*}[htbp]
    \centering
    \includegraphics[width=\textwidth]{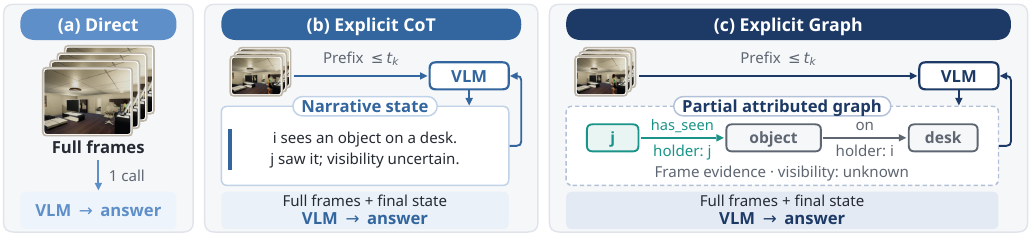}
    \vspace{-2em}
    \caption{Three inference conditions. Direct answers in one call; CoT and Graph maintain narrative or graph memory. Holder $j$ denotes $i$'s estimate of $j$'s belief. Frames and states are schematic.}
    \label{fig:inference_workflow}
    \vspace{-1em}
\end{figure*}
\noindent\textbf{Rendering and evidence measurement.}
We render static views or trajectories from agent viewpoints, aligning RGB with camera information and action records. Instance segmentation, depth, and camera geometry support instance identification and visibility measurement under field-of-view, occlusion, and interaction-state constraints. Object and event records form time-indexed histories for agents; observations preserve order and respect question cutoffs.

\noindent\textbf{Reference-answer computation.}
Answers are computed programmatically from scene geometry, executed actions, and observation histories up to the question cutoff.
Explicit perception and belief-update rules determine knowledge acquisition, retention beyond current visibility, and revision with subsequent observations, preserving uncertainty.
Cross-agent judgments relate both histories; spatial and action judgments additionally use geometry and stated goals. \looseness=-1

\noindent\textbf{Verification from the observer's perspective.}
Independently of answer computation, we verify that $i$'s observations support each inference through identifiable referents, observable actions and events, and spatial or temporal cues.
Questions and choices are grounded in recorded objects and events, then checked against answers and exported observations; unsupported candidates are revised or excluded.
We additionally conduct human verification of generated examples.
Construction protocols, measurement criteria, labeling rules, human verification details and task examples are provided in the supplementary. \looseness=-1

\section{Experiments}

\subsection{Models and Evaluation Setup}
\label{sec:experimental_setup}

\noindent\textbf{Models and inputs.}
We evaluate Qwen3.8-27B, Gemini4-Argon, GPT6-Astra, and Claude Opus 5.5 using the public task inputs.
Recordings are presented as ordered frames.

\noindent\textbf{Inference conditions.}
We compare three inference conditions (Figure~\ref{fig:inference_workflow}) to examine whether explicit memory and its organization help reconstruct others' spatial knowledge and integrate it with $i$'s own experience.
\textbf{Direct} answers from the complete public input in one call.
\textbf{Explicit CoT} maintains a natural-language reasoning summary of spatial observations and their implications for agents' knowledge.
\textbf{Explicit Graph} constructs and maintains a partial attributed scene graph from public observations, binding spatial claims to entities, belief holders, and supporting evidence.
Both memory conditions instruct the model to distinguish $i$'s information from its estimates of $j$'s beliefs and preserve uncertainty when evidence is incomplete.

For videos, CoT and Graph periodically update memory from the full available visual prefix and previous state; for static inputs, they integrate the supplied views.
Each call includes the question and task-specific references, and final answers use the complete public input and final memory.

\begin{table*}[!t]
\centering
\captionsetup{skip=3pt}
\caption{Accuracy (\%) on SpatialToM under Direct, Explicit CoT (CoT), and Explicit Graph (Graph). Macro Avg.\ is the unweighted mean over the 14 task types.}
\label{tab:spatialtom_main_results}
\begingroup
\setlength{\tabcolsep}{2.5pt}
\renewcommand{\arraystretch}{1.}
\small
\begin{tabular*}{\textwidth}{@{\extracolsep{\fill}}l*{12}{c}@{}}
\toprule
\multirow{2}{*}{Task}
& \multicolumn{3}{c}{Qwen3.8-27B}
& \multicolumn{3}{c}{Gemini4-Argon}
& \multicolumn{3}{c}{GPT6-Astra}
& \multicolumn{3}{c}{Claude Opus 5.5} \\
\cmidrule(lr){2-4}\cmidrule(lr){5-7}\cmidrule(lr){8-10}\cmidrule(lr){11-13}
& Direct & CoT & Graph & Direct & CoT & Graph
& Direct & CoT & Graph & Direct & CoT & Graph \\
\midrule
\multicolumn{13}{l}{\textbf{L0: Perception}} \\
\quad Viewpoint Inference & 36.00 & 34.00 & 36.00 & 65.00 & 65.00 & 65.00 & 53.00 & 50.00 & 50.00 & 42.00 & 47.00 & 44.00 \\
\quad Reference Frame Transformation & 34.00 & 30.00 & 48.00 & 84.00 & 94.00 & 84.00 & 77.00 & 81.00 & 81.00 & 56.00 & 60.00 & 59.00 \\
\quad Visibility Judgment & 58.90 & 46.58 & 50.68 & 82.19 & 86.30 & 83.56 & 75.34 & 69.86 & 73.97 & 53.42 & 63.01 & 63.01 \\
\quad Occlusion Reasoning & 45.45 & 48.48 & 43.94 & 71.21 & 77.27 & 69.70 & 65.15 & 59.09 & 56.06 & 48.48 & 48.48 & 46.97 \\
\quad Visibility Tracking & 41.27 & 38.10 & 38.10 & 63.49 & 61.90 & 55.56 & 57.14 & 55.56 & 50.79 & 42.86 & 46.03 & 53.97 \\
\quad Viewpoint Attribution & 42.00 & 43.00 & 42.00 & 81.00 & 89.00 & 86.00 & 77.00 & 82.00 & 74.00 & 60.00 & 60.00 & 58.00 \\
\addlinespace[2pt]
\multicolumn{13}{l}{\textbf{L1: Cognition}} \\
\quad Knowledge Attribution & 26.40 & 31.47 & 28.93 & 41.62 & 55.33 & 48.22 & 60.91 & 56.35 & 52.28 & 35.53 & 38.58 & 38.58 \\
\quad Action-Based Inference & 52.44 & 63.41 & 56.10 & 87.80 & 86.59 & 91.46 & 96.34 & 89.02 & 86.59 & 81.71 & 84.15 & 75.61 \\
\quad Belief Revision & 10.00 & 15.71 & 15.71 & 42.86 & 71.43 & 64.29 & 22.86 & 21.43 & 20.00 & 20.00 & 28.57 & 24.29 \\
\quad Spatial Correspondence & 13.74 & 15.27 & 15.27 & 62.60 & 66.41 & 61.07 & 50.38 & 37.40 & 43.51 & 30.53 & 42.75 & 37.40 \\
\quad Reference Resolution & 24.00 & 38.00 & 38.00 & 54.00 & 74.00 & 74.00 & 52.00 & 50.00 & 50.00 & 44.00 & 44.00 & 46.00 \\
\addlinespace[2pt]
\multicolumn{13}{l}{\textbf{L2: Action}} \\
\quad Route Guidance & 42.42 & 46.97 & 46.97 & 66.67 & 72.73 & 72.73 & 65.15 & 63.64 & 66.67 & 51.52 & 56.06 & 65.15 \\
\quad Information Selection & 27.59 & 27.59 & 27.59 & 37.93 & 27.59 & 37.93 & 41.38 & 62.07 & 55.17 & 27.59 & 20.69 & 27.59 \\
\quad Multi-Agent Collaboration & 57.78 & 51.11 & 46.67 & 68.89 & 75.56 & 73.33 & 71.11 & 66.67 & 68.89 & 66.67 & 64.44 & 66.67 \\
\midrule
\textbf{Macro Avg.} & 36.57 & 37.84 & 38.14 & 64.95 & 71.65 & 69.06 & 61.77 & 60.29 & 59.21 & 47.17 & 50.27 & 50.45 \\
\bottomrule
\end{tabular*}
\endgroup
\vspace{-1em}
\end{table*}

\noindent\textbf{Evaluation metrics.}
We report answer accuracy (\%) for each task type and the unweighted macro-average over all 14 types.
Single-choice answers use option matching, while set-valued answers require an exact set match.
Invalid final outputs count as incorrect.
\subsection{How Far Can Current VLMs Go in Spatial Theory of Mind?}
\label{sec:main_results}

Under Direct, performance varies substantially across tasks (Table~\ref{tab:spatialtom_main_results}), reflecting differences in reasoning demands, information density, answer spaces, and scoring criteria.
Models' understanding of spatial relations in visual observations also remains incomplete, particularly in viewpoint inference and visibility tracking.

\begin{figure*}[!t]
    \centering
    \includegraphics[width=0.93\textwidth]{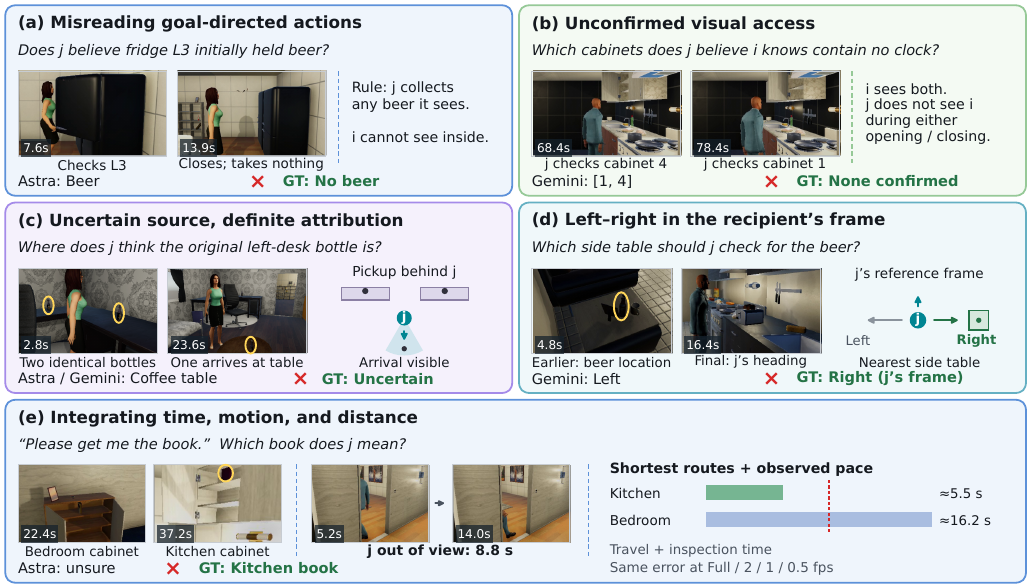}
    \captionsetup{skip=4pt}
    \caption{Representative video failures (Direct). Cropped frames show $i$'s input; wording is shortened. In (b), knowledge attribution requires $j$ to see $i$ during the check. Insets in (c,d) are schematic. Timing annotations in (e) combine shortest navigation paths, observed pace, and inspection time; they are not additional model inputs.}
    \vspace{-1em}
    \label{fig:baseline_failures}
\end{figure*}

Explicit memory can support perspective alignment while retaining errors in visual grounding.
Relative to Direct, Graph improves Reference Frame Transformation for three models and Route Guidance for all four.
The left--right error under Direct in Figure~\ref{fig:baseline_failures}(d) illustrates the alignment difficulty these tasks involve.
However, Graph lowers Occlusion Reasoning accuracy for all four models.
Inspected memory states retain visibility errors across updates, potentially affecting subsequent belief attribution.

Beyond perspective alignment, maintaining observation histories may help preserve the limits of an agent's knowledge.
CoT improves Knowledge Attribution and Belief Revision for three models, with Gemini's Belief Revision accuracy rising from 42.86\% to 71.43\%.
Figure~\ref{fig:baseline_failures}(c) illustrates the ambiguity: seeing where an object arrives does not establish which identical object was moved.
On Information Selection, Graph outperforms CoT for Gemini and Claude; only Astra improves over Direct, with CoT outperforming Graph.
CoT improves Spatial Correspondence for three models, but Astra's errors include conflating distinct objects across agents' experiences.
Because CoT and Graph also introduce additional state updates and model calls, these gains cannot be attributed to memory format alone. \looseness=-1

\findingblock{Explicit Spatial Reasoning Can Support Spatial ToM, but Remains Sensitive to Spatial Grounding Errors}{
\item VLMs recover some spatial relations and behavioral implications, but struggle to reconstruct others' knowledge.
\item Explicit memory can support perspective alignment, belief attribution, and cross-agent matching, with gains varying by model and task.
\item Visibility and identity errors can persist in memory, distorting knowledge attribution and cross-agent integration.
}

\subsection{Does Denser Visual Input Improve Spatial ToM? }
\label{sec:observation_redundancy}

We next examine whether denser visual input or explicit physical records help reconstruct agents' spatial experiences through two input studies with GPT6, using a shared cohort of 280 questions across 14 task types (20 each).

The density comparison uses the cohort's 160 recording questions across eight task types.
We compare Full reference input with uniform sampling at 2, 1, and 0.5 fps, preserving each recording's first and final frames, timestamps, and chronological order.
All three sampled conditions use the same Direct protocol; agents' physical histories and question cutoffs remain fixed.

\begin{table}[t]
\centering
\footnotesize
\renewcommand{\arraystretch}{0.78}
\setlength{\tabcolsep}{1.5pt}
\setlength{\aboverulesep}{1pt}
\setlength{\belowrulesep}{1pt}
\captionsetup{skip=2pt}
\caption{Astra Direct accuracy (\%) with reference (Full) and uniformly sampled video input (rates in fps).}
\label{tab:astra_frame_sampling}
\begin{tabular*}{\columnwidth}{@{\extracolsep{\fill}}lrrrr@{\hspace{12pt}}lrrrr@{}}
\toprule
\textbf{L1} & Full & 2 & 1 & 0.5 & \textbf{L2} & Full & 2 & 1 & 0.5 \\
\midrule
KA  & 60 & 60 & 60  & 60 & RG  & 65 & 55 & 65 & 70 \\
ABI & 95 & 90 & 100 & 90 & IS  & 45 & 25 & 20 & 40 \\
BR  & 25 & 25 & 15  & 25 & MAC & 65 & 70 & 60 & 70 \\
SC  & 50 & 50 & 35  & 35 &     &    &    &    &    \\
RR  & 50 & 55 & 50  & 50 &     &    &    &    &    \\
\midrule
\textbf{L1 avg.} & \textbf{56.0} & \textbf{56.0} & \textbf{52.0} & \textbf{52.0}
& L1--L2 & 56.9 & 53.8 & 50.6 & 55.0 \\
\bottomrule
\end{tabular*}
\vspace{-2em}
\end{table}

In this cohort, Spatial Correspondence reaches 50\% with Full or 2-fps input, compared with 35\% at 1 or 0.5 fps (Table~\ref{tab:astra_frame_sampling}).
Repeated views may help consolidate the spatial relations needed to match objects across the agents' separate explorations.
Other tasks show no consistent benefit: Knowledge Attribution is unchanged, while Information Selection performs best at the sparsest of the three sampling rates. The same five Information Selection questions incorrectly receive nonempty answers at every sampling rate.
Here, $i$ observes $j$ inspecting cabinets, but $j$ does not see $i$; the answers nevertheless treat $i$'s acquired knowledge as something $j$ believes $i$ knows.

\findingblock{Denser Views Help Build Spatial Knowledge, but Offer Limited Gains in Belief Reconstruction}{
\item In this cohort, denser views help match objects across agents' separate explorations.
\item Recurring errors still conflate what $i$ knows with what $j$ believes $i$ knows.
}

\subsection{How Much Do Oracle Physical Records Help?}
\label{sec:input_modality}

We supply textual world ground truth (GT): the initial scene graph (with object placements) and time-aligned camera poses and action histories up to the question cutoff.
These records replace videos; questions, options, and static images are retained.
For recording questions, the comparison changes both representation, from video to text, and information access, from partial views to complete physical records.
Observation and belief states remain implicit: the model must reconstruct what $i$ could observe and infer $j$'s knowledge from $i$'s perspective.
We use all 280 questions in the shared cohort, with one answer call per question in each condition and the original scoring rules.

\begin{table}[t]
\centering
\footnotesize
\renewcommand{\arraystretch}{0.78}
\setlength{\tabcolsep}{3pt}
\setlength{\aboverulesep}{1pt}
\setlength{\belowrulesep}{1pt}
\captionsetup{skip=2pt}
\caption{Astra accuracy (\%) with visual Direct input and textual world GT. Both conditions use the same 280 questions.}
\label{tab:structured_world_astra}
\begin{tabular*}{\columnwidth}{@{\extracolsep{\fill}}lrr@{\hspace{14pt}}lrr@{}}
\toprule
\textbf{L0} & Direct & World GT & \textbf{L1 / L2} & Direct & World GT \\
\midrule
VI  & 55 & 90  & KA  & 60 & 85  \\
RFT & 75 & 100 & ABI & 95 & 100 \\
VJ  & 75 & 95  & BR  & 25 & 45  \\
OR  & 65 & 85  & SC  & 50 & 90  \\
VT  & 55 & 90  & RR  & 50 & 50  \\
\cmidrule(l){4-6}
VA  & 75 & 100 & RG  & 65 & 100 \\
\cmidrule(r){1-3}
\textbf{L0--L1} & \textbf{61.8} & \textbf{84.5} & IS  & 45 & 90  \\
\textbf{L0--L2} & \textbf{61.1} & \textbf{87.1} & MAC & 65 & 100 \\
\bottomrule
\end{tabular*}
\vspace{-2em}
\end{table}

World GT raises overall accuracy from 61.1\% to 87.1\% (Table~\ref{tab:structured_world_astra}).
Accuracy improves on all Perception tasks and on Spatial Correspondence; the recurring Information Selection errors described above are corrected.

Errors remain in linking spatial entities and events to each agent's limited experience.
On Belief Revision, Astra preserves uncertainty in only five of fourteen questions requiring it; the remaining answers assign definite locations although $j$ cannot identify the moved object.
Other errors include assuming $j$ remembers an unseen book and confusing cabinet locations. Conversely, Reference Resolution reveals excessive uncertainty despite spatiotemporal constraints on j's observations. Under both Full visual input and World GT, Astra answers unsure on all 20 questions, including ten with determinate answers.

\findingblock{Textual World GT Improves Spatial ToM, but Uncertainty Handling Remains Unreliable}{
\item Physical records improve spatial judgments, knowledge attribution, and cross-agent matching.
\item Astra assigns definite beliefs under ambiguity yet answers unsure despite sufficient spatiotemporal evidence.
}

\subsection{Discussion: Do Models Spontaneously Apply Spatial ToM in Planning?}
\label{sec:qualitative_planning}

In VQA, the question itself can prompt a model to consider another agent's spatial experience.
In ordinary interaction, however, the model must recognize for itself when that experience can provide useful information.
Correct answers therefore do not establish whether a model spontaneously applies Spatial ToM to learn about the shared environment and inform cooperation.

In paired VirtualHome videos of a cooperative scenario we construct, $i$ observes two objects returned to separate cabinets, while $j$ sees a different placement in each video.
Both agents must collect one object each simultaneously without communication; a stated policy sends $j$ to the target it observed, requiring $i$ to choose the other cabinet.
Independent GPT6 probes correctly locate the targets and predict $j$'s different destinations, yet its direct plans choose the same cabinet in both videos, duplicating $j$'s destination in one. These unexecuted plans may indicate a gap between question-elicited knowledge attribution and its use in coordination.
The model can predict a partner's action from that partner's spatial experience when asked, but does not consistently use the resulting constraint to select a complementary action.

\vspace{-1em}
\paragraph{Conclusion}
SpatialToM benchmarks intersubjective embodied spatial intelligence through others' spatial knowledge. Evaluations reveal challenges connecting shared geometry with agents' limited experiences. Memory and richer evidence help, yet uncertainty persists even with world records. A planning case exposes the gap between belief attribution and coordination. Progress requires linking agents' knowledge to joint action---toward understanding and acting through one another's experiences.

\section*{Acknowledgement}
We thank 2077AI for providing computational resources in support of this research.
This work was supported by ONR MURI (N00014-24-1-2748), IITP funding from the Korean Government (MSIT) (No. RS-2024-00457882, National AI Research Lab Project), and Grant AW1134392 (Reasoning in Motion) from the TRI University 3.0 Program.
We also acknowledge computational support from Google through the Google Gemini Academic Program.

\section*{LLM Usage Statement}
We used ChatGPT, Codex, and Claude to assist with manuscript language editing and selected engineering tasks, including code implementation and debugging.
During dataset construction, we used GPT6-Astra through API and Codex-based calls to propose exploration routes and agent interactions under scene and action constraints (Sec.~\ref{sec:dataset_construction}).
These proposals were executed and checked in VirtualHome, with ground-truth labels computed from simulator records and task-specific rules.
The models and inference procedures used for benchmark evaluation are described in Sec.~\ref{sec:experimental_setup}.
The authors take full responsibility for the manuscript, implementation, dataset, and reported results.

\FloatBarrier
{
    \small
    \bibliographystyle{ieeenat_fullname}
    \bibliography{main}
}

\clearpage
\hypersetup{pageanchor=false}
\appendix
\setcounter{page}{1}
\brokenpenalty=10000

\input{sec/supp_contents}

\section{Task Descriptions and Examples}
\label{app:tasks}

SpatialToM comprises 14 tasks across Perception, Cognition, and Action. Given agent $i$'s visual observations, the model reasons about $j$'s visual access, knowledge, or behavior. Different viewpoints and exploration histories create information asymmetries. All tasks use visual question answering, including action selection and prediction.

\subsection{L0: Perception}
The Perception tier provides SpatialToM's spatial foundation: reconstructing another agent's visual perspective and access from scene geometry and pose grounds subsequent belief attribution and action reasoning. VI and RFT concern perspective transformation, VJ and OR concern object visibility, and VT and VA connect visual access and viewpoints across time.

\paragraph{Viewpoint Inference (VI).}
From $i$'s observation, the model infers how the scene appears from $j$'s position and orientation and selects the matching first-person view (Fig.~\ref{fig:examples_l0}(a)). The four candidates depict the same focal object from different camera poses, making the surrounding spatial layout essential to distinguishing $j$'s perspective.

\paragraph{Reference Frame Transformation (RFT).}
Inferring $j$'s perspective requires expressing the observed spatial relations in $j$'s reference frame. We introduce mirrored views and alternative camera poses to probe this transformation, asking the model to select the view whose left--right relations and viewing orientation agree with $j$'s actual pose (Fig.~\ref{fig:examples_l0}(b)).

\paragraph{Visibility Judgment (VJ).}
The model determines which of four objects identifiable in $i$'s image is visually accessible to $j$, accounting for $j$'s pose, field of view, and occlusion (Fig.~\ref{fig:examples_l0}(c)).

\paragraph{Occlusion Reasoning (OR).}
We arrange objects and occluders in the environment so that an object visible to $i$ is hidden from $j$ (Fig.~\ref{fig:examples_l0}(d)), requiring reasoning about how scene geometry constrains another agent's visual access. The model identifies an occluded object among visible alternatives, or a visible object among occluded and out-of-view alternatives.

\paragraph{Visibility Tracking (VT).}
VT extends VJ's single-moment judgment to tracking visual access as another agent moves and turns. From chronological observer images, the model identifies the object never visible to $j$ at any supplied moment (Fig.~\ref{fig:examples_l0}(e)). Because each distractor is visible at least once, the model must integrate visibility evidence from the full sequence.

\paragraph{Viewpoint Attribution (VA).}
Given a first-person view of $j$, the model identifies the corresponding moment among four chronological observer images (Fig.~\ref{fig:examples_l0}(f)). The candidates differ in $j$'s position, orientation, or surrounding landmarks, requiring the model to relate what $j$ sees to its pose within the scene observed by $i$.

\subsection{L1: Cognition}
The Cognition tier examines spatial belief attribution and integration. KA, ABI, and BR infer $j$'s knowledge and beliefs from its observations, actions, and subsequent evidence. SC and RR compare $i$'s beliefs with those attributed to $j$ and integrate information across their experiences, aligning object identities and spatial relations while preserving differences in knowledge and uncertainty.

\paragraph{Knowledge Attribution (KA).}
Agent $j$ explores a room, observing different objects in succession, while $i$ watches the exploration. Questions ask which objects $j$ has observed, which it remembers despite their being outside its current view (Fig.~\ref{fig:examples_l1}(a)), which candidates it has never seen, or in what order the objects were first observed. By varying encounter order and later visibility, the design separates accumulated knowledge from current perception: knowledge of previously observed objects persists even after they leave view. The model must track $j$'s visual access over time and retain the resulting knowledge, while distinguishing its observation history from $i$'s own.

\paragraph{Action-Based Inference (ABI).}
Agent $j$ aims to find and gather targets. Its stated search policy is to inspect each visited location, collect all targets found there, and close containers afterward. Interpreting these actions through the goal and policy lets $i$ infer unseen information, such as opaque container contents. Retrieving a target from a cabinet supports its initial presence there; closing it after inspection without collection supports initial absence when actions and hands remain observable (Fig.~\ref{fig:examples_l1}(b)). The model selects all supported statements about $j$'s beliefs concerning initial states, linking evidence to source locations, distinguishing $i$'s knowledge from beliefs attributed to $j$, and preserving uncertainty when evidence is insufficient.

\paragraph{Belief Revision (BR).}
An agent revises its location beliefs according to its own observations. We place two same-category objects at distinct locations; $j$ observes their initial locations before $i$ moves one. Varying $j$'s visual access to the transfer distinguishes evidence of change from evidence establishing a particular object's new location. Partial observations can reveal movement while leaving the moved object's identity or destination unresolved (Fig.~\ref{fig:examples_l1}(c)); without evidence of change, the prior location belief persists. For an object identified by its initial location, the model infers $j$'s current belief, distinguishing supported locations from partial knowledge or unresolved identity and separating $i$'s knowledge of the transfer from $j$'s conclusions.

\paragraph{Spatial Correspondence (SC).}
Integrating another agent's account with one's own spatial knowledge requires establishing whether both refer to the same physical objects. Agents $i$ and $j$ explore the rooms in turn, observing same-category objects. Agent $j$ then briefly reports the targets it observed and their relative locations (Fig.~\ref{fig:examples_l1}(d)). The model interprets $j$'s reports in light of observed evidence about its exploration route, using encounter order, room context, and supporting or containing furniture to distinguish shared encounters from discoveries outside $i$'s earlier experience. These spatial relations allow the two agents' knowledge to be combined while preserving the identities of the objects each has encountered. The model selects the option identifying all reports about objects previously encountered by $i$.

\paragraph{Reference Resolution (RR).}
When several same-category objects fit $j$'s description, $i$ uses evidence of $j$'s exploration history to resolve the ambiguity. We place two such objects in separate containers. Agent $i$ first explores these locations, then observes $j$ depart, explore partly out of view, and return with a request (Fig.~\ref{fig:examples_l1}(e)). Using $i$'s exploration history, the model combines $j$'s observed travel direction, elapsed exploration time, and distances from its observed positions to candidate locations to constrain which $j$ could have reached and inspected. Comparing this exploration range with $i$'s known candidates resolves the request only when one remains possible; the model answers \texttt{unsure} when both do.

\subsection{L2: Action}
The Action tier examines how spatial beliefs support cooperation. RG uses a partner's perspective and knowledge to guide or predict its actions, IS attributes what the partner believes the observer knows, and MAC uses the partner's predicted choice to select a complementary action.

\begin{figure*}[!t]
\centering
\includegraphics[width=\textwidth]{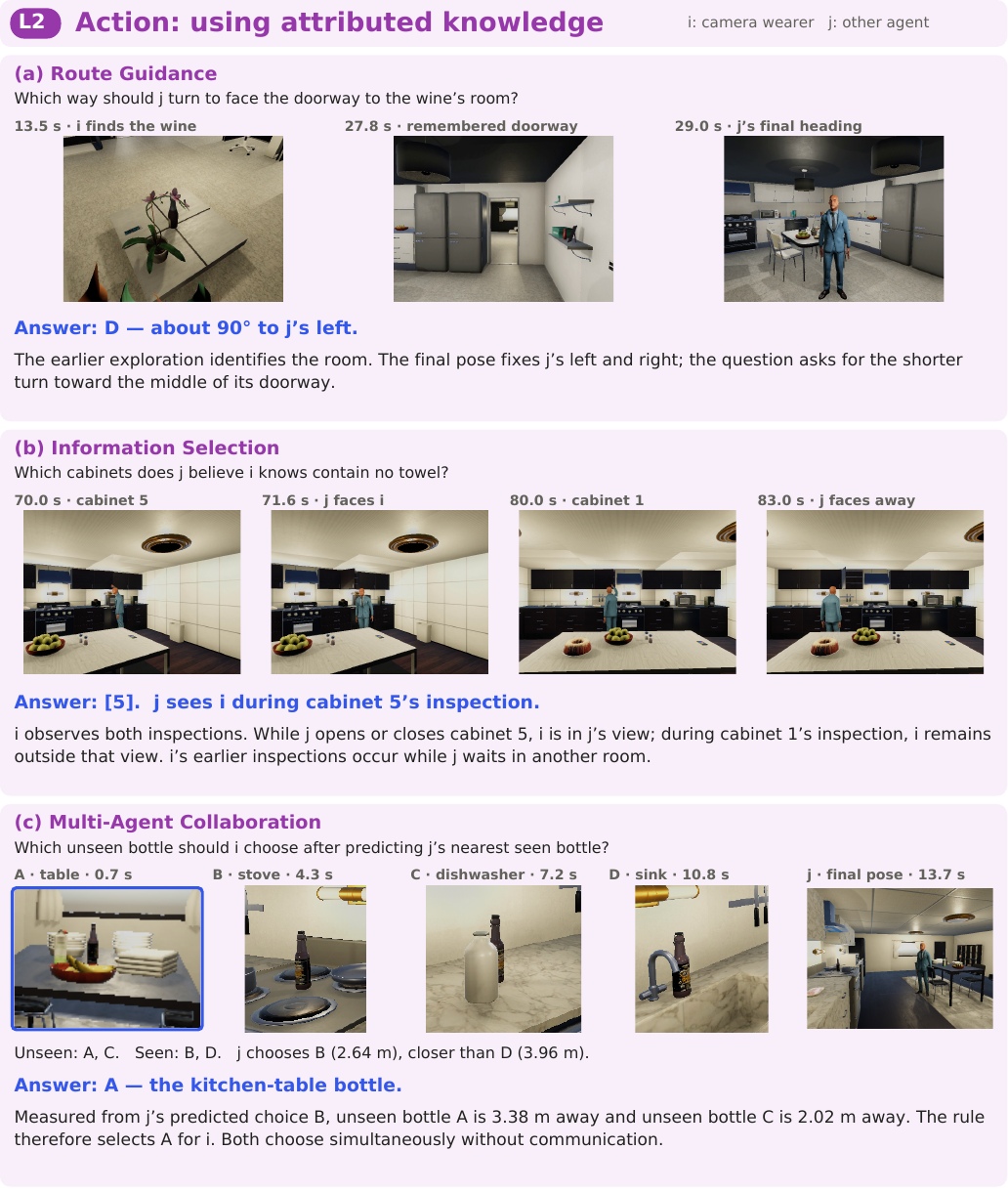}
\caption{\textbf{Action examples.} RG uses $j$'s reference frame, IS attributes what $j$ believes $i$ knows, and MAC selects an object complementary to $j$'s predicted choice. Distance values are annotations from the source records.}
\label{fig:examples_l2}
\end{figure*}

\paragraph{Route Guidance (RG).}
Guiding $j$ toward a target or predicting its actions requires combining $i$'s spatial beliefs with $j$'s goal and past experiences. Within a room, the model selects a description of the target's supporting surface by type, side, and relative distance in $j$'s reference frame, distinguishing descriptions centered on $j$ from egocentric alternatives. Across rooms, $i$ finds a target and returns; the model combines $i$'s remembered layout with $j$'s heading to select the shorter turn toward the target room's doorway (Fig.~\ref{fig:examples_l2}(a)). In a third setting, $j$ sees a target but misses its relocation by $i$ after turning away. Predicting $j$'s turn requires distinguishing its remembered target location from $i$'s updated knowledge.

\paragraph{Information Selection (IS).}
Agents $i$ and $j$ explore a series of opaque containers in turn, and $i$ must infer which containers $j$ believes $i$ knows contain no target (Fig.~\ref{fig:examples_l2}(b)). Agent $i$'s initial inspections are private, while its visibility to $j$ varies during $j$'s later inspections. Under the stated rule, $j$ attributes knowledge that a container contains no target to $i$ only when it sees $i$ during opening or closing. Private exploration and observations made outside $j$'s view can increase $i$'s knowledge without changing what $j$ believes $i$ knows. Such second-order belief reasoning is an important requirement for cooperation: understanding what a partner considers shared helps an agent decide what information still needs to be communicated and coordinate subsequent actions.

\paragraph{Multi-Agent Collaboration (MAC).}
Agents $i$ and $j$ must make complementary choices among objects distributed throughout the environment, with both deciding simultaneously and without communication. Agent $i$ knows the candidate locations and observes $j$'s exploration, while $j$ follows a stated policy of choosing the nearest object it has seen. The model uses $j$'s observation history and current position to anticipate its choice, then selects for $i$ an object unseen by $j$ and farthest from that predicted choice (Fig.~\ref{fig:examples_l2}(c)). This cooperative setting requires the model to translate a partner's spatial knowledge into a prediction of its action and use that prediction to guide its own spatial decision.

Sections~\ref{app:l0_construction}--\ref{app:l2_construction} detail the episode construction and reference-answer rules for these tasks.

\begin{figure*}[!t]
\centering
\includegraphics[width=\textwidth]{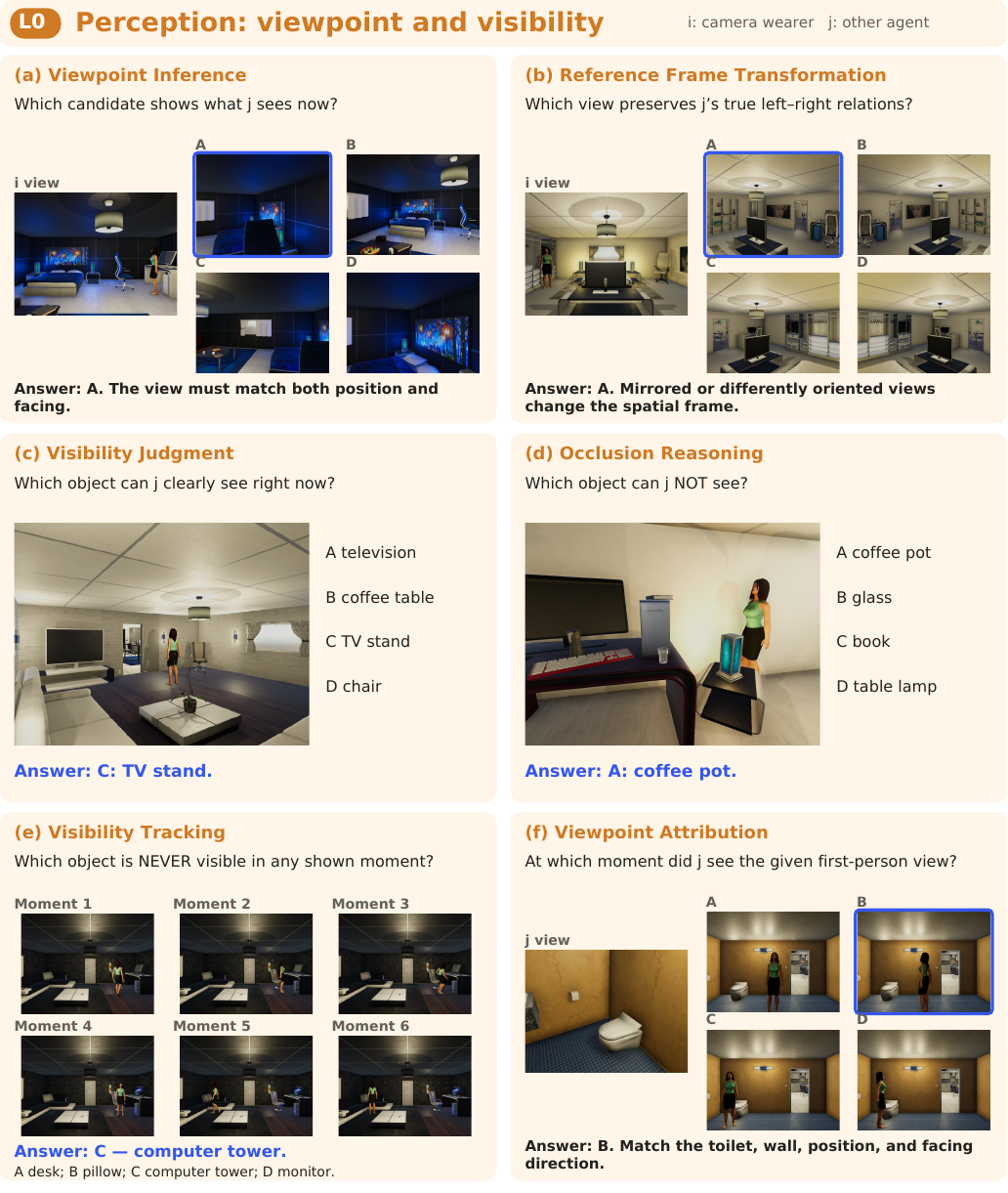}
\caption{\textbf{Perception examples.} Six tasks cover viewpoint inference, reference-frame transformation, visibility, occlusion, visibility tracking, and viewpoint attribution. VI, RFT, and VA retain all four image choices; VT retains all six supplied moments.}
\label{fig:examples_l0}
\end{figure*}

\begin{figure*}[!t]
\centering
\includegraphics[width=\textwidth]{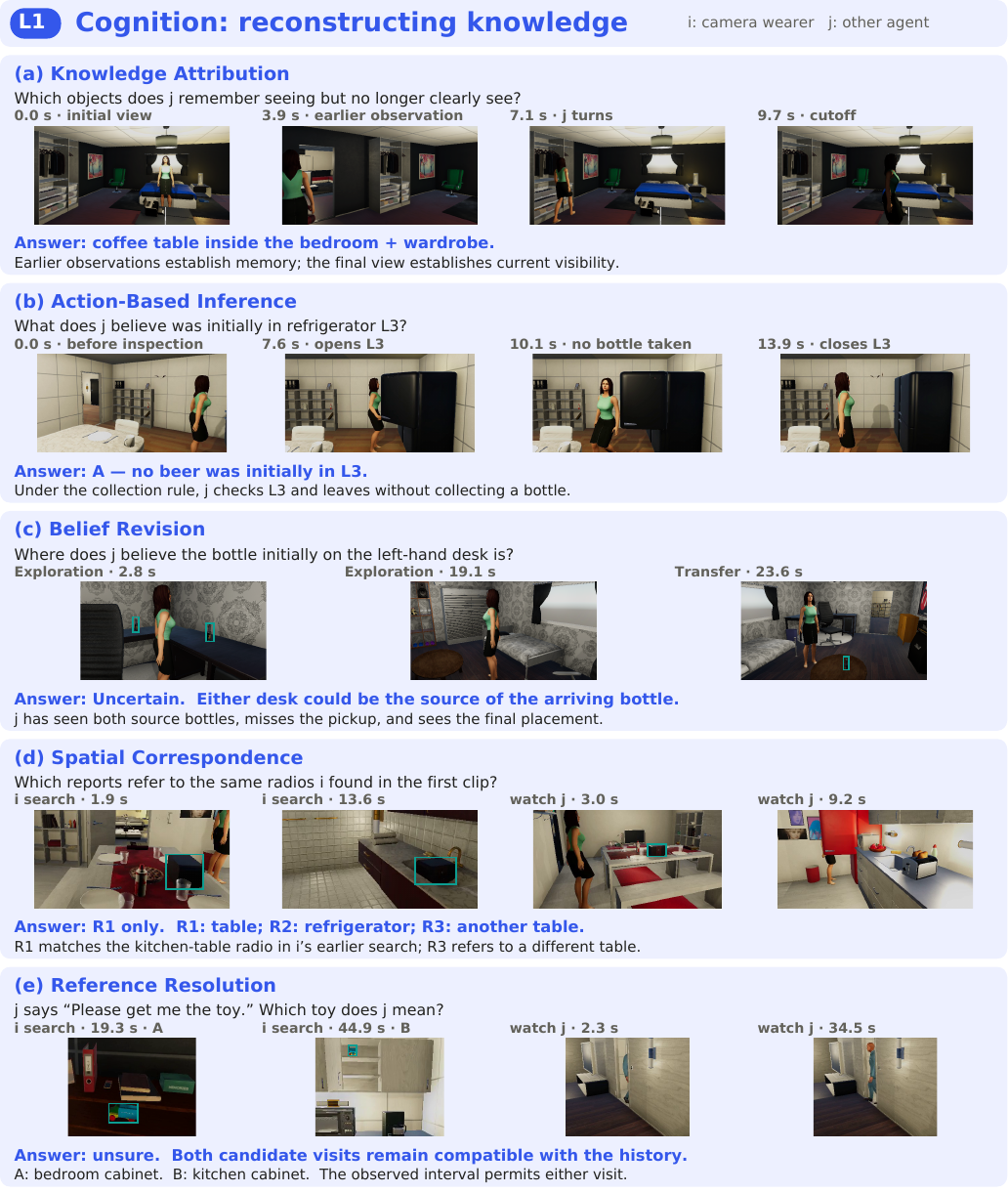}
\caption{\textbf{Cognition examples.} Observation histories and actions support knowledge attribution, action-based inference, belief revision, spatial correspondence, and reference resolution. Timestamps are local to each clip.}
\label{fig:examples_l1}
\end{figure*}

Figures~\ref{fig:examples_l0}--\ref{fig:examples_l2} show one example per task. Image choices retain their original order; video panels show selected events. Frames are cropped and questions shortened for legibility. Outlines, answer marks, and explanations are figure annotations.

\clearpage
\section{Dataset Composition and Task Formats}
\label{app:dataset_composition}

\subsection{Dataset Composition}
SpatialToM contains 1,172 questions: 502 in Perception, 530 in Cognition, and 140 in Action. Table~\ref{tab:supp_task_formats} lists the counts and formats for all 14 task types. The recording-based tasks use 812 unique video clips, with some clips shared across questions.

\begin{table}[!ht]
\centering
\small
\setlength{\tabcolsep}{4pt}
\begin{tabular}{@{}lrll@{}}
\toprule
Task & Questions & Visual input & Answer \\
\midrule
\multicolumn{4}{@{}l}{\textbf{L0: Perception (502)}} \\
VI  & 100 & Images & A--D \\
RFT & 100 & Images & A--D \\
VJ  & 73  & Images & A--D \\
OR  & 66  & Images & A--D \\
VT  & 63  & Image sequence & A--D \\
VA  & 100 & Images & A--D \\
\midrule
\multicolumn{4}{@{}l}{\textbf{L1: Cognition (530)}} \\
KA  & 197 & One recording & 1--8 \\
ABI & 82  & One recording & Set of A--D \\
BR  & 70  & Two recordings & 1--5 \\
SC  & 131 & Two recordings & 1--8 \\
RR  & 50  & Two recordings & A, B, \texttt{unsure} \\
\midrule
\multicolumn{4}{@{}l}{\textbf{L2: Action (140)}} \\
RG  & 66  & One recording & One option \\
IS  & 29  & One recording & Location set \\
MAC & 45  & One recording & A--D \\
\midrule
\textbf{Total} & \textbf{1,172} & & \\
\bottomrule
\end{tabular}
\caption{\textbf{Task counts and formats.} Abbreviations follow Section~\ref{app:tasks}. Visual inputs include task-specific reference and option images where applicable. RG has four, five, or seven options depending on its subtype.}
\label{tab:supp_task_formats}
\end{table}

\subsection{Input and Answer Formats}
Each question combines a task premise, a query, and visual observations. Reference images identify objects or locations; option images provide candidate views. Ordered images and recordings preserve the observation sequence. Questions state the agents' roles, relevant spatial conventions, and any action policy used to interpret behavior.

Most tasks return one option label. KA and SC may describe sets within their options, but the response still selects one option. ABI returns a set of statement labels, while IS returns a set of location identifiers, including the empty set. BR includes an uncertain-location option, and RR allows \texttt{unsure}. Scoring follows Section~\ref{app:experiments}.

\section{Construction Overview}
\label{app:construction}
\label{app:common_construction}

We construct the tasks in Section~\ref{app:tasks} by varying scene layouts and interaction histories to create differences in the agents' spatial knowledge.

\subsection{Scene and Object Selection}
We use VirtualHome's scene graph to select query objects, supporting surfaces or containers, and navigation landmarks. Object categories, room membership, geometry, and supported interactions determine suitable placements and agent starting poses. Simulator states and initial renders confirm the resulting layout.

\subsection{Episode Planning and Execution}
Procedural sampling and language-model proposals specify routes, viewing directions, and interactions. Model-assisted planning receives the relevant scene geometry, object identities, allowed actions, and required observation events. Proposals are executed through native navigation and interaction commands. Execution feedback guides revisions to routes, poses, or layouts; candidates that fail the task checks are excluded.

\subsection{Observation Records}
RGB observations are aligned with camera poses, instance segmentation, depth, and action records. These measurements establish object visibility and event histories for each agent. Video recordings use 10 fps at $640\times480$ resolution; model-specific input processing is described in Section~\ref{app:model_settings}. Public inputs contain $i$'s observations and task-specific reference or candidate images.

\subsection{Answer Computation and Verification}
Task-specific rules map executed states and observation histories to reference answers. Temporal questions use evidence up to their cutoff, including the frame confirming an interaction. We also check that $i$ can identify the queried objects and observe the events or spatial relations supporting the answer. The assembled question, options, and exported observations are checked for consistency. Human review is described in Section~\ref{app:human}.

\section{Perception Task Construction}
\label{app:l0_construction}

The six Perception tasks use static images or ordered image sequences, combining $j$'s pose with scene geometry to recover a view or determine visual access. Candidates vary in camera position, viewing direction, occlusion, or observation time. Recorded camera poses and visibility evidence determine the reference answers.

\subsection{Viewpoint Inference (VI)}
We select a pose of $j$ and a focal object identifiable from $i$'s view. The correct option is rendered from $j$'s recorded camera pose, while distractors change the camera position or orientation around the object. All four candidates depict the same focal object, requiring the model to use surrounding objects' relative positions and the scene layout to identify the view consistent with $j$'s pose. We check camera placement, target visibility, and differences between candidate views to ensure valid and distinguishable images.

\subsection{Reference Frame Transformation (RFT)}
We construct candidate views that differ in orientation or left--right relations, including mirrored images and views rendered from alternative camera poses. The correct view preserves $j$'s recorded position, facing direction, and left--right relations. Mirrored candidates retain scene content while reversing left and right; alternative camera poses change the viewing direction. These variations probe left--right interpretation and orientation matching in reference-frame transformation. We check that candidate views are distinguishable and exclude alternatives that closely resemble the correct view because of scene symmetry, reducing ambiguity caused by visually equivalent choices.

\subsection{Visibility Judgment (VJ)}
We select one object clearly visible to $j$ and three nonvisible alternatives, all identifiable in $i$'s image. Visibility is determined from connected instance regions in $j$'s first-person view. Nonvisible alternatives must have neither assigned nor ambiguous target pixels, avoiding cases where uncertain small visible regions are labeled as absent. Keeping all candidates identifiable to $i$ focuses the task on inferring $j$'s visual access from the observer's perspective.

\subsection{Occlusion Reasoning (OR)}
We construct object--occluder arrangements in which a target visible to $i$ can be blocked by geometry in front of $j$. Instance masks, depth, and camera geometry distinguish visible objects, objects blocked inside $j$'s viewing frustum, and objects outside it. For an occluded target, we check both its inclusion in the viewing frustum and the presence of nearer occluding geometry in its projected region, providing geometric evidence for its invisibility. Questions select either an occluded object among visible alternatives or a visible object among alternatives that include both occluded and out-of-view objects. These configurations require the model to combine viewing direction with occlusion relations when judging visibility.

\subsection{Visibility Tracking (VT)}
During $j$'s continuous exploration, we select three, four, or six chronological observer images as it moves and turns. The four candidates comprise one object visible at the final moment, two previously visible objects that are no longer visible at that moment, and one object never visible at any supplied moment, which is the answer. This configuration separates current from past visibility and requires retaining evidence from earlier observations. Each candidate is identifiable in at least one displayed observer image, and every distractor has a clearly visible occurrence in $j$'s view. Labels are computed only over the displayed moments, keeping the scope of the judgment aligned with the supplied observation sequence.

\subsection{Viewpoint Attribution (VA)}
We pair synchronized observer and first-person views at several poses of $j$. Four observer images form the chronological choices, and one corresponding first-person view becomes the query. The query's recorded pose and synchronized pairing determine the correct answer. Candidate moments have distinguishable positions, orientations, or surrounding landmarks, providing spatial evidence for matching across viewpoints. The model must relate the scene layout in the query image to $j$'s pose at each candidate moment to identify the corresponding observation.

\section{Cognition Task Construction}
\label{app:l1_construction}

The five Cognition tasks connect observation histories to retained knowledge, belief revision, and correspondence across agents' experiences. We use VirtualHome's scene graph to select objects, interaction locations, and exploration routes, then combine executed actions with synchronized observations to establish the evidence available to each agent and compute reference answers.

\subsection{Knowledge Attribution (KA)}
We select candidate objects using room membership, object categories, and geometry in VirtualHome's scene graph, and independently select navigation landmarks for route planning. The scene layout and initial observations guide $j$'s walking and turning, so that objects enter its view in different orders and leave it later in the exploration. Selecting query objects independently of navigation landmarks allows discoveries during travel. During execution, first-person observations are aligned with camera poses and instance segmentation to record when objects are observed. We check that $i$ can identify the candidates and follow the observations needed for each question.

A clear glimpse establishes an observation, and knowledge persists after the object leaves view. Let $K_j(t)$ contain objects observed by cutoff $t$, and $V_j(t)$ those currently visible. Questions query $K_j(t)$, remembered objects $K_j(t)\setminus V_j(t)$, never-seen candidates, or first-observation order. Simultaneous first observations do not define a strict order. Answers must remain supported when observations made while $j$ is unreadable to $i$ are removed.

\subsection{Action-Based Inference (ABI)}
We screen VirtualHome containers and supporting surfaces for target placement, using room membership, geometry, and supported interactions to select source locations and a separate collection destination. After configuring the targets' initial presence at these sources, we execute $j$'s search under a stated collect-all policy through native walking, opening, grabbing, placing, and closing actions. Action records link each inspection and transfer to its source, while reference images identify the queried locations. The observer sees inspections and handling actions, which can reveal contents hidden from its own view.

A completed inspection followed by retrieval or source-linked transport supports initial presence. A completed check and closing without collection supports initial absence when the relevant actions and hands remain observable. Locations without either event remain unresolved. Labels describe contents before collection and become available at the confirming event.

\subsection{Belief Revision (BR)}
We use the geometry of supporting surfaces and surrounding walkable areas to select two distinct source locations and a transfer destination, then place two same-category objects, one at each source. Agent $j$ first explores and observes both objects, then remains stationary while $i$ transfers one of them. Adjusting the layout, $j$'s viewing pose, and $i$'s transfer route controls whether $j$ sees the removal and can link the object to its destination. Removal, transport, and placement events are aligned with $j$'s observations to distinguish evidence about location change, object identity, and destination. The question identifies the object by its initial source.

\begin{center}
\small
\setlength{\tabcolsep}{3pt}
\renewcommand{\arraystretch}{1.12}
\begin{tabular}{@{}p{0.16\columnwidth}p{0.43\columnwidth}p{0.35\columnwidth}@{}}
\toprule
Removal & Later evidence & Reference belief \\
\midrule
Observed & Moved object linked to destination & New location \\
Observed & Destination unobserved & Moved; location unknown \\
Missed & Same-category arrival; source unresolved & Uncertain \\
Missed & No subsequent transfer evidence & Initial location \\
\bottomrule
\end{tabular}
\end{center}

For arrival-only episodes, the route through both sources and the pickup remain outside $j$'s view, followed by a visible placement. This provides destination evidence without identifying which object moved. We check labels against the recorded observations and verify that the final object locations agree with the executed transfer.

\subsection{Spatial Correspondence (SC)}
We use room connectivity and object--support or object--container relations in the scene graph to place same-category objects at different locations and plan routes that produce both shared observations and discoveries unique to each agent. Agent $i$ explores first, then observes $j$ explore. Observation records retain object identities, their supporting surfaces or containers, and first-encounter times. Agent $j$ truthfully reports each observed target and its supporting surface or container in first-encounter order, including discoveries during travel. Each report refers to one object. Similar descriptions at different locations require the model to use exploration routes and spatial context to identify the physical instance being described; reference answers use recorded instance identifiers to preserve object identity across observations.

A report matches if its object was observed by $i$ during its first exploration. The correct option identifies all matching reports, or indicates that none match.

\subsection{Reference Resolution (RR)}
We select two opaque containers based on room connectivity and supported interactions, then place two same-category objects, one in each. Agent $i$ first explores both locations, then observes $j$ depart, explore partly out of view, and return with a request referring to one object. Native execution records confirm visits and container interactions, while observation records capture $j$'s visible departure and return positions and the intervening unobserved interval. Labels depend on candidate locations compatible with this evidence available to $i$.

Compatibility uses scene connectivity, observed departure and return, and estimated travel and inspection time. Using $i$'s successful container-opening position as a reference, native navigation provides path lengths from the observed departure position to each candidate container and onward to the observed return position. The timing model assumes a shared exploration pace estimated from $i$'s continuous walking and combines these path lengths with calibrated opening and closing times. Timing margins allow for uncertainty in travel and interaction duration. A location remains possible if another unobserved interval permits the visit or the required timing evidence is missing. One compatible location yields its label; two yield \texttt{unsure}. The executed visit must remain compatible.

\section{Action Task Construction}
\label{app:l2_construction}

The three Action tasks use attributed knowledge to select guidance, identify shared information, or coordinate object choices. All use visual question answering. Episodes are planned from scene geometry and checked through native execution. Body turns proceed in $30^\circ$ steps recorded at their endpoints, as stated in the questions.

\subsection{Route Guidance (RG)}
We construct three episode families. Each ends with a readable pose of $j$ and requires an answer in its reference frame.

\paragraph{Within-room guidance.}
Agent $i$ finds a target on a surface while $j$ faces away. Options describe surfaces by type, side, and relative distance from $j$. An egocentric distractor describes the target from $i$'s frame but resolves to another surface from $j$'s. We check that the options identify distinct surfaces and that exactly one executed instruction reveals the target.

\paragraph{Doorway turns.}
Agent $i$ visits an adjacent room, finds the target, and returns to face $j$. The question asks for the shorter turn toward the doorway leading to that room. The answer is computed from $j$'s final heading and the doorway midpoint.

\paragraph{Previously observed locations.}
Agent $j$ sees the target on surface $A$, then turns away while $i$ moves it to $B$. We verify that the transfer remains outside $j$'s view. The answer predicts a turn toward $A$, using $j$'s retained knowledge and final pose.

\subsection{Information Selection (IS)}
We use six opaque cabinets, one containing the target. Agent $i$ privately inspects three cabinets that contain no target, then watches $j$ inspect two such cabinets. The inspected sets either overlap in one cabinet or are disjoint. The target remains unseen. Consecutive takes preserve pose and visual continuity.

The answer contains the cabinets that $j$ believes $i$ knows contain no target. A completed inspection contributes only when $j$ can see $i$ during its opening or closing actions, under the rule stated in the question. We retain cases where $i$ is clearly in view or fully out of view and exclude intermediate cases. Private inspections by $i$ contribute to its own knowledge but not to this attributed set. Responses are exact sets of cabinet identifiers, including the empty set.

\subsection{Multi-Agent Collaboration (MAC)}
Four identical bottles are identified by their supporting furniture. Agent $j$ observes exactly two, while $i$ sees their locations and witnesses $j$ acquiring that knowledge. The other two remain unobserved by $j$. Scene and trajectory checks establish these observation sets and distinguish the candidate locations.

The agents choose simultaneously without communication. Agent $j$ takes the nearest bottle it has seen; $i$ selects the bottle unseen by $j$ that is farthest from $j$'s predicted choice. With candidate set $C$, observed set $K_j$, final position $p_j$, and horizontal straight-line distance $d$,
\begin{equation}
o_j=\arg\min_{o\in K_j}d(p_j,o),\qquad
o^*=\arg\max_{o\in C\setminus K_j}d(o_j,o).
\end{equation}
Distance comparisons must remain distinct under small position perturbations. Layouts also ensure that choosing solely by distance from $j$ gives a different answer.

\section{Human Verification}
\label{app:human}

We reviewed sampled questions, observations, and reference answers from each tier during task design and refinement, with particular attention to cross-checking reference answers against key evidence frames and inspecting rendering quality.

Perception review covered agent poses, object readability, and visibility or occlusion. Cognition review checked object identity across events and consistency between attributed beliefs and the agent's observations. Action review covered reference frames, knowledge attribution, and decision-relevant distances. We also checked that viewpoint changes and recording cutoffs preserved the evidence needed to answer.

Feedback informed changes to layouts, camera views, routes, evidence checks, and question wording. We inspected new samples after revision and repeated this process before confirming the final task design and data version.

\paragraph{Pilot human baseline.}
We conducted a pilot human evaluation on 70 questions, sampling five from each of the 14 task types. Table~\ref{tab:pilot_human} compares model accuracy with human accuracy averaged across participants. No formal ethics review was sought because this was an internal pilot conducted solely by authors of this work, who consented to reporting their aggregate results.

\begin{table}[t]
    \centering
    \small
    \setlength{\tabcolsep}{5pt}
    \begin{tabular}{lccc}
        \toprule
        Model & Direct & CoT & Graph \\
        \midrule
        Qwen3.8-27B   & 42.86 & 40.00 & 54.29 \\
        Gemini4-Argon & 71.43 & 75.71 & 70.00 \\
        GPT6-Astra    & 61.43 & 60.00 & 58.57 \\
        Claude Opus 5.5& 52.86 & 60.00 & 57.14 \\
        \midrule
        Human (pilot) & \multicolumn{3}{c}{\textbf{93.57}} \\
        \bottomrule
    \end{tabular}
    \caption{Overall accuracy (\%) on the 70-question pilot subset. Model scores are taken from existing evaluations of the same questions.}
    \label{tab:pilot_human}
\end{table}

\section{Inference Protocol}
\label{app:experiments}
\label{app:memory}

\subsection{Inputs and Scoring}
Visual evaluation uses question text, reference and option images, and ordered recording frames. Simulator states and derived labels are excluded. Single-choice answers use option matching; set-valued answers require an exact set match. Invalid final outputs count as incorrect. We report per-task accuracy and the unweighted mean across all 14 types.

\subsection{Direct and Memory-Based Inference}
Direct answers from the complete input in one call. Explicit CoT and Explicit Graph first maintain a state across observations. Each update receives the question, task references, full available observation prefix, and previous state. The updated state replaces the previous one, with a soft target of at most 700 words. Both conditions add $K$ update calls before a final answer call using the complete input and final state.

Explicit CoT stores a natural-language state; Explicit Graph stores a structured spatial belief graph. Section~\ref{app:prompt_examples} gives prompt excerpts and recorded examples of both formats.

\section{Prompt and Memory Examples}
\label{app:prompt_examples}

\subsection{Shared Memory Instructions}
Memory updates combine the task question and visual evidence with instructions for tracking the agents' information. The following excerpts give the shared instructions:
\begin{quote}
\small
Distinguish information available to i from beliefs i attributes to j.\par\smallskip
Update j's belief only when evidence supports that j perceived or learned a change.\par\smallskip
Preserve older beliefs when j has not obtained the relevant update.\par\smallskip
Return a complete replacement state, retaining still-relevant previous facts.
\end{quote}

\subsection{Memory Format Instructions}
\paragraph{Explicit CoT.}
The prompt requests a natural-language state:
\begin{quote}
\small
Write a concise natural-language summary of the observed evidence, how it changes or preserves i's information and i's estimate of j's beliefs, and the resulting current understanding.\par\smallskip
Include stable entity names, supporting frame IDs, relevant spatial layout, observation times, and uncertainty where useful in the narrative.
\end{quote}

\paragraph{Explicit Graph.}
The prompt organizes the state into five fields: \texttt{entities}, \texttt{relations}, \texttt{spatial}, \texttt{updates}, and \texttt{uncertainties}. Its belief-holder convention is:
\begin{quote}
\small
\texttt{belief\_holder=i} records i's current estimate; \texttt{belief\_holder=j} records i's estimate of j's belief.
\end{quote}
For KA, a \texttt{has\_seen} relation stores remembered observations, while \texttt{visible\_to\_j\_now} records current visibility. Both methods construct a memory before the final answer call described in Section~\ref{app:experiments}.

\subsection{Recorded Memory Example}
We use the KA example in Figure~\ref{fig:examples_l1}, which asks which objects $j$ remembers seeing but can no longer clearly see. The excerpts below come from GPT6-Astra's final memory states for this recording. In the CoT state, CT denotes the bedroom coffee table and W the wardrobe.

\paragraph{CoT excerpt.}
\begin{quote}
\small
The turn and approach give a clear view of W, supported by frames 7--20; its first observation follows CT's. CT subsequently leaves j's forward view, but remains remembered. $\ldots$ The wardrobe falls behind j and remains remembered.
\end{quote}

\paragraph{Graph excerpt.}
The corresponding wardrobe relation is shown below, with field names spaced for readability:
\begin{center}
\small
\begin{tabular}{@{}ll@{}}
\toprule
Field & Recorded value \\
\midrule
Subject / predicate / object & $j$ / \texttt{has\_seen} / wardrobe \\
Belief holder & $j$ \\
Visible to $j$ now & \texttt{false} \\
Last supported frame & 97 \\
Evidence frames & 7, 14, 20, 78, 81, 97 \\
\bottomrule
\end{tabular}
\end{center}
The graph also retains the uncertainty entry: ``Wardrobe is currently behind j despite remaining visible to i.'' These excerpts show how the two formats retain an earlier observation as $j$ changes pose. Both final responses select option 6: the bedroom coffee table and wardrobe.

\section{Model Settings}
\label{app:model_settings}

\subsection{Memory Update Schedules}
GPT6-Astra, Qwen3.8-27B, and Claude Opus 5.5 use 10-second video intervals with at most five updates. For total duration $T$ accumulated across clips, $K=\min(\lceil T/10\rceil,5)$. Inputs up to 50 seconds use successive 10-second boundaries and the final partial interval; longer inputs use five evenly spaced boundaries. Visibility Tracking updates once per supplied image, while other image tasks construct one state from the supplied views.

Gemini4-Argon uses 5-second video updates with the frame grids described below. GPT6-Astra, Gemini4-Argon, and Claude Opus 5.5 use the medium reasoning setting.

\subsection{Video Input Processing}
\paragraph{Gemini4-Argon.}
Recordings are sampled at 5 fps and packed chronologically into $4\times4$ image grids, retaining the final source frame. Static references are supplied separately. Each memory update uses only frames from its available prefix.

\paragraph{Claude Opus 5.5.}
Recordings use 10 fps at $640\times480$ resolution, except Information Selection, which uses uniform 5-fps sampling. Frames are re-encoded at lower JPEG quality to fit request limits while retaining the original resolution.

\section{Additional Studies}
\label{app:additional_studies}

\subsection{Video Density}
The input studies use GPT6-Astra on a shared cohort of 280 questions, with 20 per task type. The density study uses its 160 recording questions across the eight Cognition and Action tasks. Uniform sampling at 2, 1, and 0.5 fps preserves first and final frames, timestamps, and chronological order. These conditions use Direct inference with the original questions and cutoffs; Full denotes the reference visual input.

\subsection{World Ground Truth}
All 280 questions are evaluated with textual records of the initial scene graph, object placements, camera poses, and action histories up to the cutoff. Complete physical records replace partial video observations; questions, options, and static images remain. This changes both representation and information access. Derived visibility and belief labels are omitted, so the model reconstructs agents' knowledge from the records. Each question uses one answer call and the original scoring rules.

\subsection{Qualitative Planning}
Paired recordings show $j$ observing different object placements. Independent GPT6-Astra probes ask for target locations, $j$'s predicted destination, and $i$'s plan for simultaneous collection without communication. We compare these responses to examine how attributed knowledge informs planning. Plans are assessed without simulator execution.

\let\addtocontents\SpatialToMSavedAddtocontents

\end{document}